\documentclass[10pt,twocolumn,letterpaper]{article}
\usepackage[pagenumbers]{cvpr}
\usepackage{amsmath,amssymb}
\usepackage{booktabs}
\usepackage{xspace}

\newcommand{\R}{\mathbb{R}}

\newcommand{\Conv}{\operatorname{Conv}}
\newcommand{\sigm}{\operatorname{sigmoid}}

\definecolor{cvprblue}{rgb}{0.21,0.49,0.74}
\usepackage[pagebackref,breaklinks,colorlinks,allcolors=cvprblue]{hyperref}
\def\confName{CVPR}
\def\confYear{2027}
\title{PAKAN: Pixel-Adaptive Kolmogorov--Arnold Network Modules for Pansharpening}
\author{Haoyu Zhang\thanks{Equal contribution.} \quad
Haojing Chen\footnotemark[1] \quad Zhen Zhong \quad
Liang-Jian Deng\thanks{Corresponding author.}\\
University of Electronic Science and Technology of China}

\begin{document}
\maketitle
\vspace*{-34pt}
\begin{abstract}
Most pansharpening networks process every location through the same fixed feature transformation: the nonlinearity is learned once and then applied unchanged across edges, textures, and homogeneous regions, even though these regions need different amounts of injected detail. We present PAKAN, two pixel-adaptive Kolmogorov--Arnold network modules that replace this transformation. PAKAN expands features in a fixed spline basis and modulates the learned edge functions with two gates derived from the input---a per-location gate shared over feature channels and a per-channel gate pooled over space---and the two branch responses are multiplied. The same two modules enter two-stream hosts through a fusion interface and single-stream hosts through a refinement interface. On U2Net, PanNet, and FusionNet over WorldView-3, QuickBird, and GaoFen-2, replacing the original unit improves PSNR in all nine sensor--backbone pairs ($0.121$--$1.440$\,dB) and lowers SAM and ERGAS in all nine, with full-resolution HQNR gains in all nine, at $8.6$--$13.0\%$ extra parameters and $10.6$--$12.8\%$ extra FLOPs. A component study shows that the spline response and the input-conditioned gates both contribute: removing either lowers PSNR and raises SAM and ERGAS on every sensor, and the full module is best on all three metrics.
\end{abstract}

\section{Introduction}
\label{sec:introduction}

Pansharpening fuses a high-resolution panchromatic (PAN) image with a low-resolution multispectral (MS) image to recover a high-resolution multispectral output~\cite{wald1997fusion,vivone2020benchmark}. A deep network for this task learns a feature transformation and applies it at every location. The nonlinear part of that transformation is fixed after training: a roof edge, a vegetation boundary, and a homogeneous field are all mapped through the same function, even though each needs a different amount of injected detail. The fusion process is itself nonlinear in a way that differs from place to place: the relation between a surface and its recorded spectrum changes between shadowed facades, open water, and dense canopy, and so does the amount of detail to inject. A single global nonlinearity cannot express this variation.

Prior work has made the kernel input-dependent, made the activation learnable, or made a prescribed activation input-conditional; none of these combines a learned spline basis with coefficients regenerated from the input. Adaptive-convolution operators condition the kernel on local content---LAGConv adds a global harmonic bias~\cite{jin2022lagconv}, LGPConv perturbs a shared kernel~\cite{zhao2023lgpconv}, and CANConv partitions pixels by non-local similarity~\cite{duan2024canconv}---but the activation applied after the kernel remains fixed. Kolmogorov--Arnold layers replace the fixed activation with a spline basis whose coefficients are learned~\cite{liu2024kan}, but those coefficients are learned once and then frozen, so every location is still mapped by the same function. Dynamic ReLU is a closely related prior: it predicts the parameters of a piecewise-linear activation from the input~\cite{chen2020dyrelu}, but its basis is prescribed rather than learned, so the input can only select among fixed hinge components rather than re-weight a learned spline basis. To our knowledge, PAKAN is the first work to bring pixel-adaptive Kolmogorov--Arnold networks to pansharpening.

\begin{figure*}[t]
  \centering
  \includegraphics[width=0.97\textwidth]{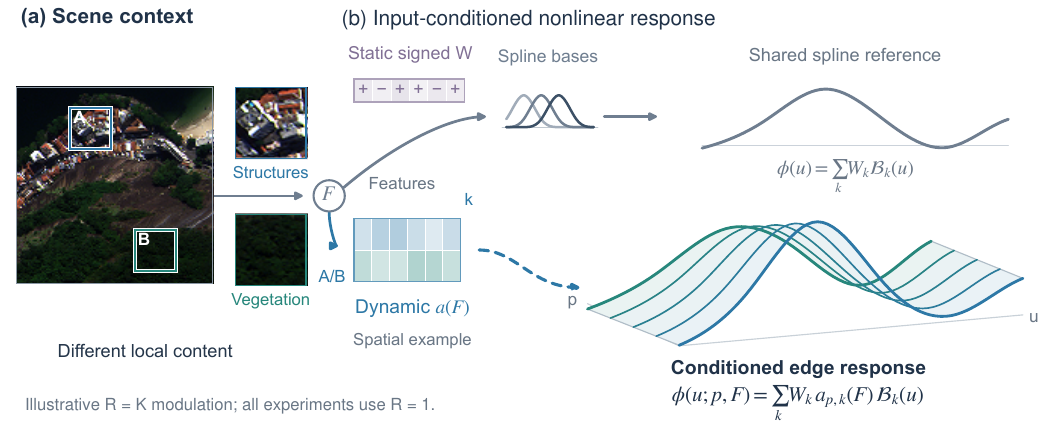}
  \caption{Input-conditioned spline responses. (a) A remote-sensing scene and its crops illustrate local context. (b) Input features supply shared basis evaluation and dynamic modulation, while signed learned weights remain fixed at inference. The surface is drawn for the illustrative $R=K$ setting; all reported experiments use $R=1$, where a single gate rescales the shared edge function. Context colors are illustrative. Coefficients and curves are schematic, not measurements from a trained model.}
  \label{fig:overview}
\end{figure*}

PAKAN implements this combination in two modules that replace the feature-processing unit of an existing host. Each branch keeps one fixed spline basis whose signed coefficients are learned and shared across the image; two gates---a per-location gate shared over channels and a per-channel gate pooled over space---modulate those edge functions before aggregation, and the two branch responses are multiplied. Through a fusion interface for two-stream hosts and a refinement interface for single-stream hosts, the two modules replace the original unit in U2Net, PanNet, and FusionNet. Figure~\ref{fig:overview} illustrates the response core.

Across WV3, QB, and GF2, all nine sensor--backbone pairs improve in PSNR by $0.121$--$1.440$\,dB and lower SAM and ERGAS; full-resolution HQNR improves on all nine; the cost is $8.6$--$13.0\%$ parameters and $10.6$--$12.8\%$ FLOPs. A component study shows that the spline response and the input-derived gates both contribute: removing either lowers PSNR and raises SAM and ERGAS, and the full module is best on every sensor.

Our contributions are:
\begin{itemize}
  \item The first pixel-adaptive Kolmogorov--Arnold modules for pansharpening: they modulate the edge functions of a fixed spline basis rather than re-weighting features or kernels, with a per-location gate shared over channels and a per-channel gate pooled over space that generate $B\,R\,(H_fW_f+C)$ modulation entries instead of a dense $B\,H_fW_f\,C\,R$.
  \item Two interfaces---fusion ($2$-to-$1$) and refinement ($1$-to-$1$)---that insert the same response core into three host networks; all nine sensor--backbone pairs improve in PSNR ($0.121$--$1.440$\,dB) and lower SAM and ERGAS, with full-resolution HQNR gains in all nine pairs and $8.6$--$13.0\%$ parameter overhead.
  \item A component study showing that the spline response and the input-conditioned gates both contribute: removing either lowers PSNR and raises SAM and ERGAS, and the full module achieves the best results.
\end{itemize}

\section{Related Work}
\label{sec:related}

\paragraph{Deep pansharpening.}
PNN learns an image-to-image reconstruction mapping~\cite{masi2016pnn}. PanNet uses high-pass-domain processing~\cite{yang2017pannet}, while DiCNN and FusionNet learn detail-injection mappings~\cite{he2019dicnn,deng2020fusionnet}. U2Net maintains spatial and spectral feature streams and integrates them at multiple scales~\cite{peng2023u2net}. Pan-Mamba introduces state-space processing for cross-modal interaction~\cite{zhou2024panmamba}. Adaptive operators have also been developed specifically for this task: LAGConv uses local-context kernels and a global bias~\cite{jin2022lagconv}, LGPConv introduces learnable Gaussian perturbations~\cite{zhao2023lgpconv}, and CANConv combines content-adaptive convolution with non-local similarity~\cite{duan2024canconv}. PAKAN contributes a spline-based feature operator that can be incorporated into existing reconstruction pipelines. This operator-level objective motivates comparisons with the host networks in addition to comparisons with complete methods.

\paragraph{Conditional feature transformations.}
SE and CBAM condition channel and spatial feature weighting on the input~\cite{hu2018senet,woo2018cbam}. Dynamic convolution conditions a mixture of convolution kernels~\cite{chen2020dynamic}, and Dynamic ReLU conditions piecewise-linear activation parameters, including variants with different spatial and channel sharing patterns~\cite{chen2020dyrelu}. Conditionally parameterized convolutions and feature-wise affine modulation extend the same idea to kernels and feature statistics~\cite{yang2019condconv,perez2018film}. Pixel-adaptive convolution predicts a spatially varying kernel from local input features~\cite{su2019pac}. These approaches establish that both transformation parameters and activation parameters can depend on image content. PAKAN instead predicts a spatially varying gate on the coefficients of a shared spline basis. A gate shared across basis functions rescales a learned edge function; a gate varying over the basis index changes the relative basis contributions. The two regimes are defined in Sec.~\ref{sec:conditional}.

\paragraph{Learnable spline transformations.}
KANs use learnable univariate functions to parameterize feature transformations~\cite{liu2024kan}. In a spline implementation, the basis supplies nonlinear responses and the learned coefficients determine their combination. KAN 2.0 extends the formulation toward scientific discovery~\cite{liu2024kan2}, and a matched-compute comparison finds that MLPs generally outperform KANs except for symbolic formula representation, tracing the advantage there to the B-spline activation~\cite{yu2024kanfair}. We therefore use KAN as a spline parameterization, not as a claim of superior function approximation. PAKAN combines this parameterization with input-dependent modulation. We distinguish the parameters learned across the training set from the modulation computed for an individual input. This distinction identifies what is shared by the trained model and what changes with location or channel context.

\section{Method}
\label{sec:method}

\subsection{Task and Feature Representation}
Let $\mathbf{P}\in\R^{B\times1\times H\times W}$ be the PAN image and $\mathbf{M}\in\R^{B\times C_{\mathrm{MS}}\times H/r\times W/r}$ the MS image, with spatial ratio $r$. A backbone predicts $\widehat{\mathbf{I}}\in\R^{B\times C_{\mathrm{MS}}\times H\times W}$. PAKAN operates on intermediate features rather than directly on the measured spectral bands. We write an input feature tensor as $\mathbf{F}\in\R^{B\times C\times H_f\times W_f}$, and use $p$ to index a feature-map location. Batch, input-channel, output-channel, and basis indices are denoted by $b,i,j,k$, respectively.

\subsection{Spline Responses with Conditioned Coefficients}
\label{sec:conditional}
For a scalar feature value $u$, a KAN-style edge function can be written as
\begin{equation}
  \phi_{ji}(u)=\sum_{k=1}^{K}W_{ji,k}\,\mathcal{B}_{k}(u),
  \label{eq:static}
\end{equation}
where $\mathcal{B}_{k}$ is a spline basis function and $W_{ji,k}$ is a learned coefficient. A layer aggregates these univariate responses over input channels. In the static form, $W$ is shared across locations and images at inference.

PAKAN introduces modulation $a_{b,p,i,k}(\mathbf{F})$ computed from the input, giving the conditional spline response
\begin{equation}
  Y_{b,p,j}=\sum_{i=1}^{C}\sum_{k=1}^{K}
  W_{ji,k}\,a_{b,p,i,k}(\mathbf{F})\,
  \mathcal{B}_{k}(F_{b,p,i}).
  \label{eq:conditional}
\end{equation}
The effective coefficient is $\widetilde W_{b,p,j,i,k}=W_{ji,k}a_{b,p,i,k}(\mathbf{F})$. This separates the shared learned parameter $W$ from the input-dependent quantity $a$. Because the gate multiplies the signed coefficient $W$, a nonnegative gate leaves the sign of the effective coefficient unchanged. The sigmoid gates used for U2Net are nonnegative; the SiLU channel gate used for PanNet and FusionNet is not, so there the response can change sign. If a base or linear path is included, its response is added to Eq.~\eqref{eq:conditional}. Channel grouping restricts the sum over $i$ to the input channels connected to output $j$.

In all experiments $\mathcal{B}_k$ is a fixed power-triangular basis with $K=G$. For grid size $G$, the centers are $c_k=-1+2(k-1)/(G-1)$ and
\begin{equation}
  \mathcal{B}_k(u)=
  \frac{\bigl[\max(0,\,1-|u-c_k|/\delta)\bigr]^{q}}
       {\sum_{k'}\bigl[\max(0,\,1-|u-c_{k'}|/\delta)\bigr]^{q}},
  \qquad \delta=\frac{2}{G-1},
  \label{eq:basis}
\end{equation}
with exponent $q$. We use $(G,q)=(5,3)$ for U2Net, $(6,4)$ for PanNet, and $(4,3)$ for FusionNet. The centers and width are fixed; only the coefficients are learned. A linear path with coefficients $V_{ji}$ and a bias $\beta_j$ is added to Eq.~\eqref{eq:conditional}.

There are two useful modulation granularities. A gate shared over $k$ rescales a learned edge function, since $\sum_k W_{ji,k}a_{b,p,i}\mathcal{B}_k(u)=a_{b,p,i}\phi_{ji}(u)$. A gate that varies over $k$ can change the relative basis coefficients and hence the shape of the conditional edge function. We use the conditional-transformation terminology to distinguish these cases from unrestricted synthesis of a new activation function. The full module also includes channel aggregation and the combination of two nonlinear branches.

\subsection{Spatial and Channel Conditioning}
\label{sec:branches}
We construct two operators, $\Psi_{\mathrm{spa}}$ and $\Psi_{\mathrm{chn}}$, using Eq.~\eqref{eq:conditional} with different sharing patterns. Each has its own learned spline parameters and conditioning generator. Let $R$ be the number of generated modulation entries per unit: $R=1$ for a gate shared over the basis axis, or $R=K$ for basis-dependent modulation. This notation makes the generated tensor sizes explicit. All experiments in this paper use $R=1$.

\paragraph{Spatial conditioning.}
A local convolutional generator retains the feature-map resolution:
\begin{equation}
  \mathbf{A}_{\mathrm{spa}}
  =\sigm\!\left(\Conv_2\bigl(\eta(\Conv_1(\mathbf{F}))\bigr)\right),
  \label{eq:spa_generator}
\end{equation}
where $\eta$ denotes the intermediate pointwise nonlinearity and $\mathbf{A}_{\mathrm{spa}}\in\R^{B\times R\times H_f\times W_f}$. Its entries vary with $p$ and are broadcast over input and output channels. Thus
\begin{equation}
  a^{\mathrm{spa}}_{b,p,i,k}(\mathbf{F})
  =A^{\mathrm{spa}}_{b,k,p},
  \label{eq:spa_sharing}
\end{equation}
with the third index running over the $R$ modulation entries and constant when $R=1$. The learned $W^{\mathrm{spa}}_{ji,k}$ still depends on the connected channel pair; it is the generated modulation, not the entire edge coefficient, that is shared over channels.

\begin{figure*}[t]
  \centering
  \includegraphics[width=0.98\textwidth]{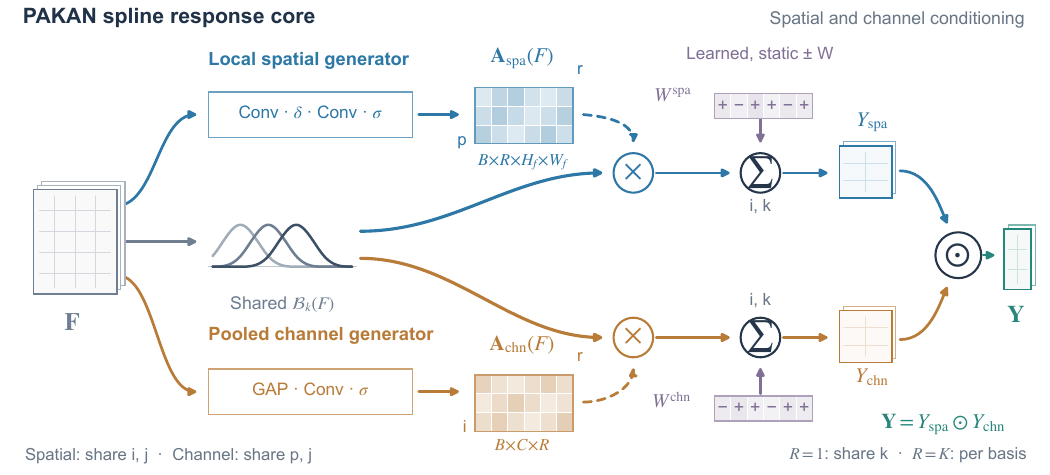}
  \caption{PAKAN response core. Input $\mathbf F$ supplies both conditioning generators and shared basis evaluation. Spatial modulation varies with $p$ and is shared over $i,j$; channel modulation varies with $i$ and is shared over $p,j$. Each branch aggregates modulated basis responses using its own signed $W$ before summing over $i,k$. Their outputs interact elementwise. Dashed ribbons carry modulation. $R=1$ shares a gate over $k$; $R=K$ modulates each basis. Base paths are omitted.}
  \label{fig:generators}
\end{figure*}

\paragraph{Channel conditioning.}
The channel generator aggregates spatial context before producing modulation:
\begin{equation}
  z_{b,i}=\frac{1}{H_fW_f}\sum_p F_{b,p,i},\qquad
  \mathbf{A}_{\mathrm{chn}}=\psi\!\left(\Conv(\mathbf z)\right).
  \label{eq:chn_generator}
\end{equation}
where $\psi$ is a pointwise activation; we use sigmoid in the U2Net configuration and SiLU in the PanNet and FusionNet configurations.
The output is reshaped to $\R^{B\times C\times R}$ and broadcast over locations and connected output channels:
\begin{equation}
  a^{\mathrm{chn}}_{b,p,i,k}(\mathbf{F})
  =A^{\mathrm{chn}}_{b,i,k}.
  \label{eq:chn_sharing}
\end{equation}
This branch conditions the contribution of input feature channels on pooled context. These channels are learned representations and need not correspond one-to-one to physical spectral bands. The two sharing patterns mirror the task: detail restoration varies with location, so the gate changes with $p$ while the operator stays identical across channels, and spectral fidelity is a scene property, so the channel gate is pooled over space.

Together they provide spatially varying and channel-dependent control without generating a dense coefficient tensor for every channel pair and location. Their modulation outputs contain $B R(H_fW_f+C)$ entries, compared with $B H_fW_f C R$ for dense modulation at every input channel and position. This is an output-tensor comparison; the generator parameter counts and the cost of spline evaluation are included in the totals in Table~\ref{tab:efficiency}. Figure~\ref{fig:generators} shows the conditioning pathways.

\subsection{Response Combination and Module Interfaces}
Both operators produce the same output shape, and their responses are combined elementwise:
\begin{equation}
  \mathcal{T}(\mathbf F)
  =\Psi_{\mathrm{spa}}(\mathbf F)\odot
   \Psi_{\mathrm{chn}}(\mathbf F).
  \label{eq:coupling}
\end{equation}
The product introduces an interaction between the two nonlinear responses at each output location and channel. Its empirical contribution is assessed against additive and concatenation-based combinations. Because the responses can be signed, the product is treated as a learned interaction rather than assumed to be a positive attention mask.

\paragraph{Single-stream refinement (1to1).}
\label{sec:interfaces}
An input $\mathbf X$ is projected to an internal feature representation by $E$, processed by the two branches, and projected to the required output width by $Q$:
\begin{equation}
  \mathcal{P}_{\mathrm{ref}}(\mathbf X)
  =Q\!\left(\mathcal{T}(E(\mathbf X))\right).
  \label{eq:refine}
\end{equation}
This interface applies to a mixed feature stream inside a residual reconstruction network. It preserves the feature-map resolution and can preserve the input channel width through the output projection.

\begin{figure*}[t]
  \centering
  \includegraphics[width=0.98\textwidth]{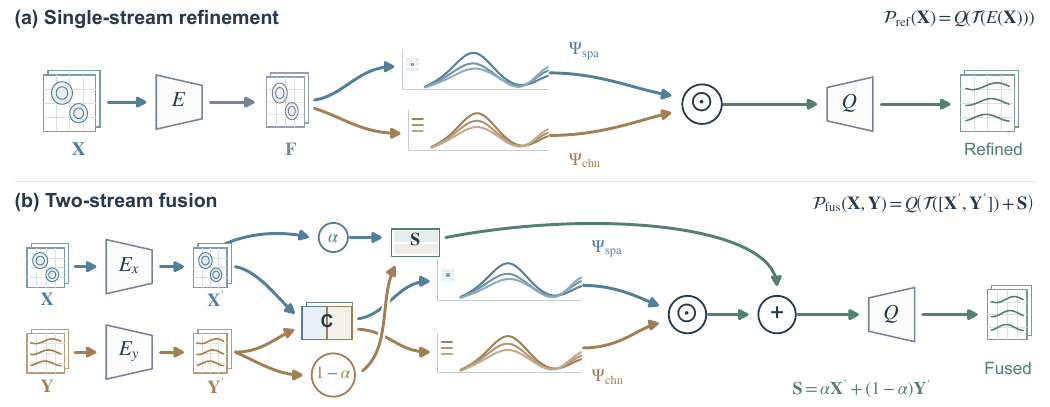}
  \caption{Module interfaces. (a) Single-stream refinement uses $Q(\mathcal T(E(\mathbf X)))$. (b) Fusion concatenates projected inputs and adds $\mathbf S=\alpha\mathbf X'+(1-\alpha)\mathbf Y'$ after the response product and before $Q$. Both interfaces combine spatial (blue) and channel (amber) responses elementwise. $E,Q$ denote feature projections and $C$ concatenation. Features and curves are schematic.}
  \label{fig:modules}
\end{figure*}

\paragraph{Two-stream fusion (2to1).}
For aligned inputs $\mathbf X$ and $\mathbf Y$, modality projections produce $\mathbf X'=E_x(\mathbf X)$ and $\mathbf Y'=E_y(\mathbf Y)$ of width $D$. Concatenation supplies the joint input to both branches:
\begin{equation}
  \mathbf U=[\mathbf X',\mathbf Y'],\qquad
  \mathcal{P}_{\mathrm{fus}}(\mathbf X,\mathbf Y)
  =Q\!\left(\mathcal{T}(\mathbf U)+\mathbf S\right),
  \label{eq:fusion}
\end{equation}
where each branch maps $2D$ input channels to $D$ output channels. A residual mixture can be parameterized as $\mathbf S=\alpha\mathbf X'+(1-\alpha)\mathbf Y'$, with $\alpha=\sigm(\gamma)$ and learned scalar $\gamma$. The shortcut carries projected input information alongside the conditioned response. Figure~\ref{fig:modules} shows the two interfaces.

\paragraph{Backbone integration and training.}
U2Net's two-stream fusion structure uses the fusion interface at five scales with channel widths $32,64,128,64,32$. The single-stream interface supports feature processing in PanNet and FusionNet after the observed inputs have been mixed, with one block in PanNet and four blocks in FusionNet. The interfaces retain the host network's reconstruction output and spatial data flow. They replace feature-processing units, rather than requiring a new end-to-end backbone. The training objective is the mean absolute reconstruction error,
\begin{equation}
  \mathcal L=\frac{1}{|\Omega|}
  \sum_{n\in\Omega}\left|\widehat I_n-I_n^{\mathrm{GT}}\right|,
  \label{eq:loss}
\end{equation}
where $\Omega$ indexes batch elements, spatial locations, and MS bands.

\section{Experiments}
\label{sec:experiments}

\subsection{Evaluation Setup}
We evaluate two questions: whether PAKAN improves different host networks, and how its learned spline processing and conditional components contribute to the observed gains. The experiments use U2Net, PanNet, and FusionNet as host networks, together with comparisons to representative complete pansharpening methods.

\paragraph{Data and protocols.}
WV3, QB, and GF2 are drawn from PanCollection~\cite{pancollection}. The spatial ratio is four. WV3 has eight MS bands; QB and GF2 have four. Each sensor provides separate training, validation, and test splits. Reduced-resolution (RR) testing uses 20 reference samples per sensor following the degradation-based evaluation protocol~\cite{wald1997fusion,vivone2020benchmark}. We report PSNR, spectral angle mapper (SAM), relative dimensionless global error in synthesis (ERGAS), and Q8 or Q4. Full-resolution (FR) evaluation uses spectral distortion $D_\lambda$, spatial distortion $D_s$, and the hybrid quality with no reference (HQNR). Lower SAM, ERGAS, $D_\lambda$, and $D_s$ indicate less error or distortion; higher PSNR and Q indicate better reconstruction. The RR and FR results provide complementary assessments and are interpreted within their respective protocols.

\paragraph{Training and inference.}
Training uses PyTorch on an RTX 4090, with $16\times16$ LR-MS patches and $64\times64$ PAN and reference patches. Inputs are normalized to $[0,1]$ using the sensor's peak (2047 for WV3/QB, 1023 for GF2) before being passed to the network. We use Adam~\cite{kingma2014adam} with batch size 32, initial learning rate $4\times10^{-4}$, and $(\beta_1,\beta_2)=(0.9,0.999)$ for 500 epochs. The learning rate is multiplied by $0.7$ every 100 epochs. Training uses pre-extracted patches without additional geometric or photometric augmentation. FR inference uses $64\times64$ HR tiles, the corresponding $16\times16$ MS tiles, and overlapping center-region stitching with padding four.

The default configuration is grid size $G=5$ with order three for U2Net, $G=6$ with order four for PanNet, and $G=4$ with order three for FusionNet, giving $K=G$ basis functions in each case. The sensitivity study varies $G$ and the order within the U2Net configuration.

\subsection{Consistent Gains across Backbones}
\label{sec:paired}
\begin{table}[t]
  \centering
  \setlength{\tabcolsep}{1.2mm}
  \small
  \caption{Reduced-resolution results. * denotes the PAKAN-equipped backbone; bold marks the PAKAN-equipped rows.}
  \label{tab:plug}
  \begin{tabular}{lcccc}
    \toprule
    \textbf{Method} & PSNR$\uparrow$ & SAM$\downarrow$ & ERGAS$\downarrow$ & Q$\uparrow$ \\
    \midrule
    \multicolumn{5}{c}{\textit{WV3 (Q8)}} \\
    \midrule
    PanNet & 37.346 & 3.613 & 2.664 & 0.891 \\
    PanNet* & \textbf{37.962} & \textbf{3.590} & \textbf{2.642} & \textbf{0.894} \\
    FusionNet & 38.047 & 3.324 & 2.465 & 0.904 \\
    FusionNet* & \textbf{38.528} & \textbf{3.210} & \textbf{2.382} & \textbf{0.911} \\
    U2Net & 39.117 & 2.888 & 2.150 & 0.920 \\
    U2Net* & \textbf{39.238} & \textbf{2.856} & \textbf{2.119} & \textbf{0.921} \\
    \midrule
    \multicolumn{5}{c}{\textit{QB (Q4)}} \\
    \midrule
    PanNet & 34.678 & 5.767 & 5.859 & 0.885 \\
    PanNet* & \textbf{36.118} & \textbf{5.438} & \textbf{5.231} & \textbf{0.892} \\
    FusionNet & 37.540 & 4.904 & 4.156 & 0.925 \\
    FusionNet* & \textbf{37.912} & \textbf{4.760} & \textbf{3.890} & \textbf{0.928} \\
    U2Net & 38.065 & 4.642 & 3.987 & 0.931 \\
    U2Net* & \textbf{38.520} & \textbf{4.439} & \textbf{3.698} & \textbf{0.939} \\
    \midrule
    \multicolumn{5}{c}{\textit{GF2 (Q4)}} \\
    \midrule
    PanNet & 40.243 & 0.997 & 0.919 & 0.967 \\
    PanNet* & \textbf{41.331} & \textbf{0.922} & \textbf{0.857} & \textbf{0.969} \\
    FusionNet & 39.638 & 0.974 & 0.988 & 0.964 \\
    FusionNet* & \textbf{40.317} & \textbf{0.933} & \textbf{0.912} & \textbf{0.966} \\
    U2Net & 43.379 & 0.714 & 0.632 & 0.982 \\
    U2Net* & \textbf{43.888} & \textbf{0.668} & \textbf{0.595} & \textbf{0.987} \\
    \bottomrule
  \end{tabular}
\end{table}

Table~\ref{tab:plug} evaluates PAKAN on three host networks. Every sensor--backbone pair improves in PSNR, with gains of $0.121$--$1.440$\,dB. PanNet gains $0.616$, $1.440$, and $1.088$\,dB on WV3, QB, and GF2, respectively. The stronger U2Net also benefits on every sensor, gaining $0.121$, $0.455$, and $0.509$\,dB. The consistency across residual and multi-scale architectures shows that the operator's utility extends beyond one host design.

The gains jointly improve reconstruction fidelity and spectral agreement: SAM and ERGAS decrease in all nine pairs. For GF2 U2Net, the $0.509$\,dB PSNR gain accompanies SAM decreasing from $0.714$ to $0.668$ and ERGAS from $0.632$ to $0.595$. These changes make the benefit visible across error measures, alongside the smaller absolute changes in Q8/Q4.

\begin{table}[t]
  \centering
  \setlength{\tabcolsep}{1.2mm}
  \small
  \caption{Full-resolution results. * denotes the PAKAN-equipped backbone; bold marks the PAKAN-equipped rows.}
  \label{tab:plug_fr}
  \begin{tabular}{lccc}
    \toprule
    \textbf{Method} & $D_\lambda\!\downarrow$ & $D_s\!\downarrow$ & HQNR$\uparrow$ \\
    \midrule
    \multicolumn{4}{c}{\textit{WV3}} \\
    \midrule
    PanNet & 0.017 & 0.047 & 0.937 \\
    PanNet* & \textbf{0.016} & \textbf{0.046} & \textbf{0.939} \\
    FusionNet & 0.024 & 0.036 & 0.941 \\
    FusionNet* & \textbf{0.021} & \textbf{0.032} & \textbf{0.947} \\
    U2Net & 0.018 & 0.031 & 0.951 \\
    U2Net* & \textbf{0.018} & \textbf{0.025} & \textbf{0.956} \\
    \midrule
    \multicolumn{4}{c}{\textit{QB}} \\
    \midrule
    PanNet & 0.043 & 0.114 & 0.849 \\
    PanNet* & \textbf{0.039} & \textbf{0.112} & \textbf{0.853} \\
    FusionNet & 0.057 & 0.052 & 0.894 \\
    FusionNet* & \textbf{0.051} & \textbf{0.037} & \textbf{0.914} \\
    U2Net & 0.059 & 0.026 & 0.916 \\
    U2Net* & \textbf{0.058} & \textbf{0.025} & \textbf{0.919} \\
    \midrule
    \multicolumn{4}{c}{\textit{GF2}} \\
    \midrule
    PanNet & 0.018 & 0.080 & 0.904 \\
    PanNet* & \textbf{0.017} & \textbf{0.079} & \textbf{0.905} \\
    FusionNet & 0.035 & 0.101 & 0.867 \\
    FusionNet* & \textbf{0.033} & \textbf{0.093} & \textbf{0.877} \\
    U2Net & 0.024 & 0.051 & 0.927 \\
    U2Net* & \textbf{0.021} & \textbf{0.048} & \textbf{0.932} \\
    \bottomrule
  \end{tabular}
\end{table}

Full-resolution evaluation reinforces the U2Net comparison: $D_s$ decreases on all three sensors, and $D_\lambda$ is unchanged at the printed precision on WV3 (Table~\ref{tab:plug_fr}). HQNR improves from $0.951$ to $0.956$ on WV3, $0.927$ to $0.932$ on GF2, and $0.916$ to $0.919$ on QB.

\subsection{Comparison with Representative Methods}
\begin{table*}[t]
  \centering
  \setlength{\tabcolsep}{1.2mm}
  \small
  \caption{Reduced-resolution comparison with the selected pansharpening methods on WV3, QB, and GF2. Best: \textbf{bold}; second: \underline{underline}. $^{*}$ denotes the PAKAN-equipped backbone.}
  \label{tab:sota}
  \begin{tabular}{lcccccccccccc}
    \toprule
    & \multicolumn{4}{c}{\textbf{WV3}} & \multicolumn{4}{c}{\textbf{QB}} & \multicolumn{4}{c}{\textbf{GF2}} \\
    \cmidrule(lr){2-5}\cmidrule(lr){6-9}\cmidrule(lr){10-13}
    \textbf{Method} & PSNR$\uparrow$ & SAM$\downarrow$ & ERGAS$\downarrow$ & Q8$\uparrow$ & PSNR$\uparrow$ & SAM$\downarrow$ & ERGAS$\downarrow$ & Q4$\uparrow$ & PSNR$\uparrow$ & SAM$\downarrow$ & ERGAS$\downarrow$ & Q4$\uparrow$ \\
    \midrule
    MTF-GLP-FS & 32.963 & 5.316 & 4.700 & 0.833 & 32.709 & 7.792 & 7.373 & 0.835 & 35.540 & 1.655 & 1.589 & 0.897 \\
    BDSD-PC & 32.970 & 5.428 & 4.697 & 0.829 & 32.550 & 8.085 & 7.513 & 0.831 & 35.180 & 1.681 & 1.667 & 0.892 \\
    TV & 32.381 & 5.692 & 4.855 & 0.795 & 32.136 & 7.510 & 7.690 & 0.821 & 35.237 & 1.911 & 1.737 & 0.907 \\
    \midrule
    PNN & 37.313 & 3.677 & 2.681 & 0.893 & 36.942 & 5.181 & 4.468 & 0.918 & 39.071 & 1.048 & 1.057 & 0.960 \\
    PanNet & 37.346 & 3.613 & 2.664 & 0.891 & 34.678 & 5.767 & 5.859 & 0.885 & 40.243 & 0.997 & 0.919 & 0.967 \\
    DiCNN & 37.390 & 3.592 & 2.672 & 0.900 & 35.781 & 5.367 & 5.133 & 0.904 & 38.906 & 1.053 & 1.081 & 0.959 \\
    FusionNet & 38.047 & 3.324 & 2.465 & 0.904 & 37.540 & 4.904 & 4.156 & 0.925 & 39.638 & 0.974 & 0.988 & 0.964 \\
    LAGConv & 38.592 & 3.103 & 2.291 & 0.910 & \underline{38.209} & 4.870 & \underline{3.812} & \underline{0.934} & 42.735 & 0.786 & 0.687 & 0.980 \\
    LGPConv & 38.147 & 3.270 & 2.422 & 0.902 & 36.443 & 4.954 & 4.777 & 0.915 & 41.843 & 0.845 & 0.765 & 0.976 \\
    Pan-Mamba & 39.012 & 2.913 & 2.184 & \underline{0.920} & 37.356 & \underline{4.625} & 4.277 & 0.929 & 42.906 & 0.743 & 0.684 & \underline{0.982} \\
    U2Net & \underline{39.117} & \underline{2.888} & \underline{2.150} & \underline{0.920} & 38.065 & 4.642 & 3.987 & 0.931 & \underline{43.379} & \underline{0.714} & \underline{0.632} & \underline{0.982} \\
    \textbf{U2Net$^{*}$} & \textbf{39.238} & \textbf{2.856} & \textbf{2.119} & \textbf{0.921} & \textbf{38.520} & \textbf{4.439} & \textbf{3.698} & \textbf{0.939} & \textbf{43.888} & \textbf{0.668} & \textbf{0.595} & \textbf{0.987} \\
    \bottomrule
  \end{tabular}
\end{table*}

Table~\ref{tab:sota} compares U2Net$^{*}$ with traditional methods (MTF-GLP-FS, BDSD-PC, and TV) and learning-based methods (PNN, PanNet, DiCNN, FusionNet, LAGConv, LGPConv, Pan-Mamba, and U2Net)~\cite{aiazzi2006mtf,vivone2019bdsd,palsson2014tv,masi2016pnn,yang2017pannet,he2019dicnn,deng2020fusionnet,jin2022lagconv,zhao2023lgpconv,zhou2024panmamba,peng2023u2net}. U2Net$^{*}$ has the highest PSNR and lowest SAM and ERGAS among the selected methods in this table on each sensor. The WV3 PSNR gain over U2Net is $0.121$\,dB, whereas the QB and GF2 gains are $0.455$ and $0.509$\,dB. The magnitude of the improvement is therefore sensor-dependent, and the host-network comparisons in Sec.~\ref{sec:paired} show that the same direction of change holds across three architectures.

\begin{figure*}[t]
  \centering
  \includegraphics[width=0.84\textwidth]{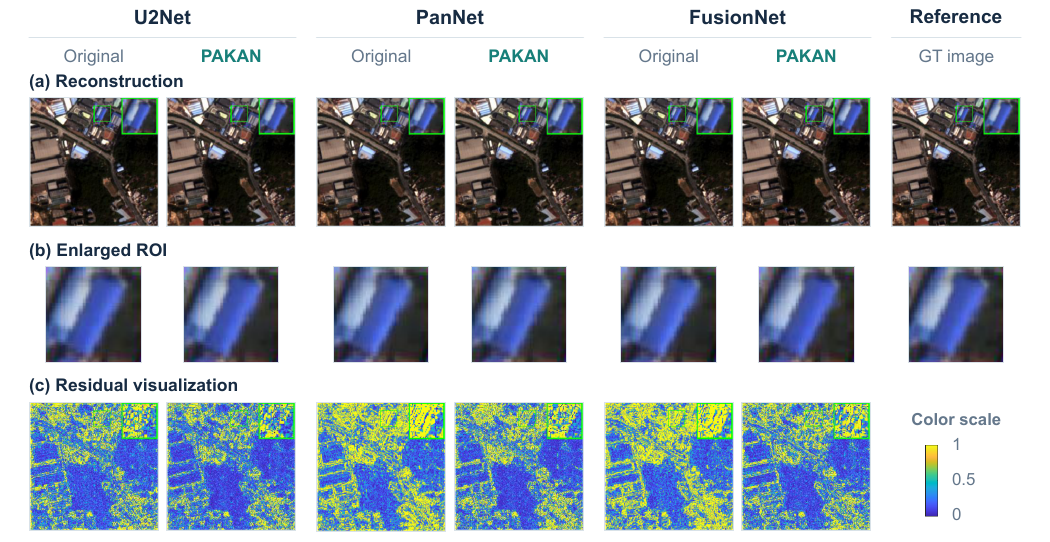}
  \caption{Reduced-resolution comparisons on a WorldView-3 scene. (a) Original and PAKAN reconstructions with the shared reference. (b) Enlarged regions from the insets in (a). (c) Residual (reconstruction minus reference) maps, contrast-stretched for display.}
  \label{fig:visual}
\end{figure*}

\subsection{Component and Design Ablations}
\begin{table*}[t]
  \centering
  \small
  \setlength{\tabcolsep}{1.2mm}
  \caption{Component and branch ablations with U2Net. Pixel-adaptive weighting (PA) uses conditional processing without KAN; static KAN uses the spline component without pixel-adaptive conditioning. Channel-only and spatial-only retain the corresponding adaptive branch.}
  \label{tab:abl_components}
  \begin{tabular}{lcccccccccccc}
    \toprule
    & \multicolumn{4}{c}{\textbf{WV3}} & \multicolumn{4}{c}{\textbf{QB}} & \multicolumn{4}{c}{\textbf{GF2}} \\
    \cmidrule(lr){2-5}\cmidrule(lr){6-9}\cmidrule(lr){10-13}
    Variant & PSNR$\uparrow$ & SAM$\downarrow$ & ERGAS$\downarrow$ & Q8$\uparrow$ & PSNR$\uparrow$ & SAM$\downarrow$ & ERGAS$\downarrow$ & Q4$\uparrow$ & PSNR$\uparrow$ & SAM$\downarrow$ & ERGAS$\downarrow$ & Q4$\uparrow$ \\
    \midrule
    Backbone & 39.117 & 2.888 & 2.150 & 0.920 & 38.065 & 4.642 & 3.987 & 0.931 & 43.379 & 0.714 & 0.632 & 0.982 \\
    PA only & 39.174 & 2.867 & 2.135 & 0.921 & 38.246 & 4.505 & 3.773 & 0.935 & 43.546 & 0.693 & 0.615 & 0.984 \\
    Static KAN & 39.191 & 2.877 & 2.144 & 0.921 & 38.342 & 4.592 & 3.889 & 0.937 & 43.774 & 0.696 & 0.625 & 0.985 \\
    Channel only & 39.162 & 2.876 & 2.141 & 0.920 & 38.217 & 4.556 & 3.812 & 0.934 & 43.578 & 0.699 & 0.620 & 0.984 \\
    Spatial only & 39.198 & 2.865 & 2.129 & 0.921 & 38.371 & 4.487 & 3.744 & 0.937 & 43.756 & 0.684 & 0.607 & 0.985 \\
    \textbf{PAKAN} & \textbf{39.238} & \textbf{2.856} & \textbf{2.119} & \textbf{0.921} & \textbf{38.520} & \textbf{4.439} & \textbf{3.698} & \textbf{0.939} & \textbf{43.888} & \textbf{0.668} & \textbf{0.595} & \textbf{0.987} \\
    \bottomrule
  \end{tabular}
\end{table*}

\paragraph{Spline processing and conditioning.}
Table~\ref{tab:abl_components} removes one part of the module at a time. A static-basis variant keeps the learned spline and removes all conditioning, and a conditioning-only variant keeps the gates and removes the spline response. Each variant alone is worse than the full module in PSNR, SAM, and ERGAS on every sensor, while the full module is best on all three. Conditioning the learned spline is therefore part of the design rather than an optional add-on: both components contribute to the observed gains.

\paragraph{Spatial and channel branches.}
The branch ablation in Table~\ref{tab:abl_components} shows that both branches improve over the backbone and that their combination produces the best evaluated results. Here the channel-only and spatial-only variants keep one adaptive branch (with its spline response) and drop the other. The spatial-only variant gives higher PSNR than the channel-only variant on all three sensors: $39.198$ versus $39.162$\,dB on WV3, $38.371$ versus $38.217$\,dB on QB, and $43.756$ versus $43.578$\,dB on GF2. Both branches are retained because the combined design improves on either one alone.

\begin{table}[t]
  \centering
  \setlength{\tabcolsep}{1.2mm}
  \small
  \caption{Attention alternatives and response-combination variants on WV3 with the U2Net backbone.}
  \label{tab:attn}
  \begin{tabular}{lcccc}
    \toprule
    \textbf{Variant} & PSNR$\uparrow$ & SAM$\downarrow$ & ERGAS$\downarrow$ & Q8$\uparrow$ \\
    \midrule
    U2Net (baseline) & 39.117 & 2.888 & 2.150 & 0.920 \\
    + Channel Attn (SE) & 39.142 & 2.875 & 2.141 & 0.920 \\
    + CBAM & 39.156 & 2.869 & 2.137 & 0.920 \\
    \midrule
    PAKAN (additive) & 39.198 & 2.871 & 2.130 & 0.920 \\
    PAKAN (concat) & 39.207 & 2.868 & 2.127 & 0.920 \\
    \textbf{PAKAN (multiplicative)} & \textbf{39.238} & \textbf{2.856} & \textbf{2.119} & \textbf{0.921} \\
    \bottomrule
  \end{tabular}
\end{table}

\paragraph{Attention alternatives and combination.}
On WV3 with U2Net, the SE and CBAM variants reach $39.142$ and $39.156$\,dB, respectively, compared with $39.238$\,dB for PAKAN (Table~\ref{tab:attn}). The multiplicative combination reaches $39.238$\,dB, compared with $39.198$\,dB for addition and $39.207$\,dB for concatenation. The accompanying SAM and ERGAS changes favor the same configuration. These results support multiplicative combination and the complete module over the tested alternatives in this setting.

\begin{table}[t]
  \centering
  \setlength{\tabcolsep}{1.5mm}
  \small
  \caption{Spline grid size and order on WV3 with U2Net$^{*}$ (the PAKAN-equipped U2Net). Default: $G=5$, order three.}
  \label{tab:hparam}
  \begin{tabular}{c@{\hskip 3pt}c@{\hskip 3pt}c@{\hskip 3pt}c@{\hskip 8pt}c@{\hskip 3pt}c@{\hskip 3pt}c@{\hskip 3pt}c}
    \toprule
    \multicolumn{4}{c}{\textbf{Spline Grid Size} $G$} & \multicolumn{4}{c}{\textbf{Spline Order}} \\
    $G$ & PSNR$\uparrow$ & SAM$\downarrow$ & ERGAS$\downarrow$ & Order & PSNR$\uparrow$ & SAM$\downarrow$ & ERGAS$\downarrow$ \\
    \midrule
    3 & 39.195 & 2.873 & 2.131 & 2 & 39.205 & 2.870 & 2.128 \\
    5 & \textbf{39.238} & \textbf{2.856} & \textbf{2.119} & 3 & \textbf{39.238} & \textbf{2.856} & \textbf{2.119} \\
    7 & 39.231 & 2.860 & 2.123 & 4 & 39.229 & 2.862 & 2.124 \\
    \bottomrule
  \end{tabular}
\end{table}

\paragraph{Spline settings.}
Table~\ref{tab:hparam} varies $G\in\{3,5,7\}$ and order in $\{2,3,4\}$ on WV3 as a post-hoc sensitivity check; the spline settings were fixed before training and are not selected by this sweep. The default, $G=5$ with order three, is best among these configurations, and every swept configuration still exceeds the U2Net backbone, the weakest by $0.078$\,dB. The setting therefore changes the size of the gain by at most $0.043$\,dB rather than creating it. PanNet and FusionNet use their own per-host settings.

\subsection{Efficiency, Repeatability, and Visual Results}
\begin{table}[t]
  \centering
  \small
  \caption{Computational cost before and after adding PAKAN. The inference-time increment is reported in the text for a $64\times64$ PAN tile on an RTX 4090.}
  \label{tab:efficiency}
  \begin{tabular}{lcc}
    \toprule
    Backbone & Parameters (M) & FLOPs (G) \\
    \midrule
    PanNet & $0.23\rightarrow0.26$ & $18.7\rightarrow21.1$ \\
    FusionNet & $1.18\rightarrow1.30$ & $43.2\rightarrow47.8$ \\
    U2Net & $1.52\rightarrow1.65$ & $52.4\rightarrow58.1$ \\
    \bottomrule
  \end{tabular}
\end{table}

\paragraph{Computational cost.}
Table~\ref{tab:efficiency} reports parameter and FLOP counts. Added parameters range from approximately $8.6\%$ for U2Net to $13.0\%$ for PanNet, while the FLOP increases are approximately $10.6$--$12.8\%$. The reported inference-time increment is $1.7$--$3.1$\,ms per $64\times64$ PAN tile on an RTX 4090.
\paragraph{Repeated runs.}
The three-run WV3 U2Net evaluation reports PSNR $39.238\pm0.005$\,dB for U2Net$^{*}$ and $39.117\pm0.006$\,dB for U2Net. SAM is $2.856\pm0.005$ and $2.888\pm0.006$, respectively. The run-to-run spread ($0.005$--$0.006$\,dB) is an order of magnitude smaller than the observed PSNR gains, so the improvements are not explained by training noise alone.

\paragraph{Qualitative comparisons.}
Figure~\ref{fig:visual} compares each backbone and its PAKAN-equipped version on the same reduced-resolution scene. The shared reference, enlarged region, and residual visualizations allow direct inspection of the two reconstructions alongside the aggregate metrics.

\subsection{Scope and Limitations}
The present study trains one model per sensor. Extending the conditioning to a shared spectral prior, so that a single module transfers across sensors, is a natural next step. The generators are separable along space and channels; a joint parameterization and direct measurement of the conditional responses would show how far the same principle extends.

\section{Conclusion}
We presented PAKAN, two pixel-adaptive Kolmogorov--Arnold modules that replace the fixed feature transformation of a pansharpening network with a fixed spline basis whose learned edge functions are modulated by an input-derived spatial gate and channel gate. Through fusion and refinement interfaces, the two modules improve all nine sensor--backbone pairs in PSNR and lower SAM and ERGAS, at $8.6$--$13.0\%$ additional parameters and $10.6$--$12.8\%$ additional FLOPs. A component study shows that both parts contribute: removing either the spline response or the input-derived gates lowers PSNR and raises SAM and ERGAS, and the full module is best on PSNR, SAM, and ERGAS.

The same interfaces apply to other multi-scale fusion backbones. Our training is sensor-specific, so whether the learned gates transfer across sensors remains untested.
\label{end:main}
\section*{LLM Usage Statement}
{\small
OpenAI GPT, DeepSeek, and Anthropic Claude assisted with writing and editing, related work, and code implementation, provided reviewer-style feedback, and generated draft figures; all research ideas, experiments, data, and conclusions are the authors' own.
}

{\small
\setlength{\bibsep}{-1.5pt}
\bibliographystyle{ieeenat_fullname}
\bibliography{references}
}
\end{document}